\documentclass[letterpaper]{article} 
\usepackage{aaai25}  
\usepackage{times}  
\usepackage{helvet}  
\usepackage{courier}  
\usepackage[hyphens]{url}  
\usepackage{graphicx} 
\usepackage{natbib}  
\usepackage{caption} 
\usepackage{algorithm}
\usepackage{algorithmic}
\usepackage{amsmath}
\usepackage{amsthm}
\usepackage[utf8]{inputenc} 
\usepackage{amsmath}
\usepackage{amssymb} 
\usepackage{multirow}
\theoremstyle{definition}
\newtheorem{definition}{\bf Definition}

\usepackage{booktabs}
\usepackage{newfloat}
\usepackage{listings}
\DeclareCaptionStyle{ruled}{labelfont=normalfont,labelsep=colon,strut=off} 
\floatstyle{ruled}
\newfloat{listing}{tb}{lst}{}
\floatname{listing}{Listing}
\title{Multi-Granularity Open Intent Classification via \\Adaptive Granular-Ball Decision Boundary}
\author{
    Yanhua Li\textsuperscript{\rm 1,2},
    Xiaocao Ouyang\textsuperscript{\rm 1,2},
    Chaofan Pan\textsuperscript{\rm 1,2},
    Jie Zhang\textsuperscript{\rm 1,2},
    Sen Zhao\textsuperscript{\rm 3},\\
    Shuyin Xia\textsuperscript{\rm 3},
    Xin Yang\textsuperscript{\rm 1,2}\thanks{Corresponding author},
    Guoyin Wang\textsuperscript{\rm 4},
   Tianrui Li\textsuperscript{\rm 5}
}
\affiliations{
    \textsuperscript{\rm 1}School of Computing and Artificial Intelligence, Southwestern University of Finance and Economics, Chengdu~611130, China\\
    \textsuperscript{\rm 2}Engineering Research Center of Intelligent Finance, Ministry of Education, Chengdu~611130, China\\
    \textsuperscript{\rm 3}Key Laboratory of Big Data Intelligent Computing, Chongqing Key Laboratory of Computational Intelligence, Chongqing University of Posts and Telecommunications, Chongqing~400065, China\\
    \textsuperscript{\rm 4}National Center for Applied Mathematics in Chongqing, Chongqing Normal University, Chongqing 401331, China\\
    \textsuperscript{\rm 5}School of Computing and Artificial Intelligence, Southwest Jiaotong University, Chengdu~611756, China\\
    yyhhliyanhua@163.com, oyxiaocao@swufe.edu.cn, pan.chaofan@foxmail.com, jiezhanglx@gmail.com, \\zhaosen@cqupt.edu.cn, xiasy@cqupt.edu.cn, yangxin@swufe.edu.cn, wanggy@cqnu.edu.cn, trli@swjtu.edu.cn

}

\usepackage{bibentry}

\begin{document}

\maketitle

\begin{abstract}
Open intent classification is critical for the development of dialogue systems, aiming to accurately classify known intents into their corresponding classes while identifying unknown intents. Prior boundary-based methods assumed known intents fit within compact spherical regions, focusing on coarse-grained representation and precise spherical decision boundaries. However, these assumptions are often violated in practical scenarios, making it difficult to distinguish known intent classes from unknowns using a single spherical boundary.  To tackle these issues, we propose a \textbf{M}ulti-granularity \textbf{O}pen intent classification method via adaptive \textbf{G}ranular-\textbf{B}all decision boundary (MOGB). Our MOGB method consists of two modules: representation learning and decision boundary acquiring. To effectively represent the intent distribution, we design a hierarchical representation learning method. This involves iteratively alternating between adaptive granular-ball clustering and nearest sub-centroid classification to capture fine-grained semantic structures within known intent classes. Furthermore, multi-granularity decision boundaries are constructed for open intent classification by employing granular-balls with varying centroids and radii. Extensive experiments conducted on three public datasets demonstrate the effectiveness of our proposed method. 
\end{abstract}
 \begin{links}
     \link{Code}{https://github.com/Liyanhuaa/MOGB}
 \end{links}
 
\section{Introduction}
The identification of user intent is critical for dialogue systems, especially as individuals increasingly rely on these systems to support both daily activities and professional tasks.
In real-world applications, dialogue systems often encounter intents from unknown (open) classes that they have not seen during training, resulting in their inability to handle these intents effectively~\cite{zheng2020out}. To address this, open intent classification aims to classify known intents while identifying unknown intents~\cite{shu2017doc}.
\begin{figure}[t]
\centering
\includegraphics[width=1\columnwidth]{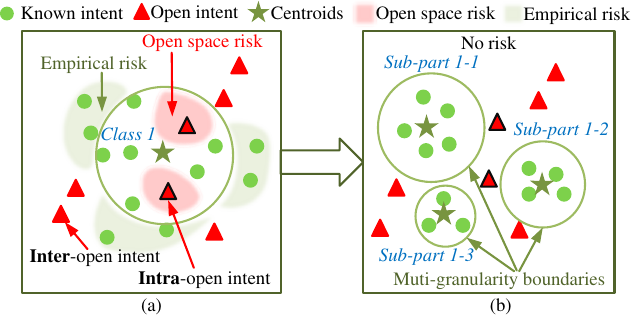} %
\caption{ (a) Previous boundary-based methods struggle to learn a boundary for open intent classification by balancing open space risk and empirical risk.
(b) Our proposed Multi-granularity decision boundary can effectively eliminate both two risks. }
\label{fig: motivation}
\end{figure}

Recent researches endeavor to develop effective methods to tackle the issue of open intent classification. 
Early methods~\cite{jain2014multi,hendrycks2022baseline} focused on establishing a $K$-class classifier and determining whether a sample belongs to unknowns by checking if its maximum probability exceeds a certain threshold. 
Due to the overfitting of deep learning approaches~\cite{guo2017calibration}, $K$-class classifiers tend to overconfidence about known intent classes and are prone to misclassifying unknown intents as known ones. To overcome this limitation, some studies train a $K$+1 classifier directly, leveraging pseudo-unknown data to represent the open intents~\cite{zheng2020out,cheng2022learning}. 

In recent years, boundary-based methods ~\cite{zhang2021deep,zhang2023learning,yang2022three} have been proposed to learn specific decision boundaries between known intents and unknowns.
These methods often involve two stages, i.e. representation learning and boundary learning. 
They mainly focused on learning more accurate decision boundaries to balance open space risk (classify unknowns as known) and empirical risk (classify known as unknowns). 
Initially, Zhang et al.~\cite{zhang2021deep} acquired feature inputs through softmax-based representation learning and obtained adaptive decision boundary (ADB) for each known intent class. Following this work, Liu et al.~\cite{liu2023effective} designed a K-center contrastive learning algorithm to learn discriminative intent representations and proposed adjustable boundary learning by expanding or shrinking the initial decision boundary. Zhang et al.~\cite{zhang2023learning} learned distance-aware intent representations and tight adaptive decision boundaries to improve open intent detection.

Existing boundary-based methods~\cite{zhang2021deep,zhang2023learning} implicitly assumed the features of known intents distributed in a compact spherical region, while intents lying outside this distribution are regarded as open intents.
It means that open intents do not exist within the distribution of a single known intent class and only appear between different known intent classes. 
However, as illustrated in Figure~\ref{fig: motivation} (a), in real-world situations, known intents do not consistently adhere to a compact spherical distribution and open intents can exist between or within the distribution of known intents (named inter-open intents and intra-open intents). 
The investigation of Gaussian Hypothesis Testing concerning known intents distribution, as presented by Zhou et al.~\cite{zhou2022knn}, revealed that merely 57\% of the known intents within the CLINC-FULL dataset conform to the Gaussian assumption. This finding further implies that the assumption of a spherical feature distribution is unreasonable.
When dealing with non-spherical intent distributions, it is difficult to distinguish known intents from unknowns using just one decision boundary. 
Furthermore, the open space risk resulting from intra-open intents can never be eliminated, regardless of how to adjust the boundary, as shown in Figure~\ref{fig: motivation} (a).

In this paper, we propose a new straightforward but powerful solution: exploring the patterns within each known intent class, dividing intents into multiple sub-parts, and establishing decision boundaries for each sub-part, as illustrated in Figure~\ref{fig: motivation} (b). 
This solution offers two benefits over earlier boundary-based methods.
On the one hand, having several sub-parts can more accurately reflect the actual distribution of known intent classes. On the other hand, it creates fine-grained decision boundaries that can more efficiently exclude intra-open intents, reducing the open space risk.

In light of this, the following three challenges come to this paper: (1) How to divide each known intent class into multiple sub-parts that accurately reflect the intent distribution? 
(2) How to learn discriminative representations that capture different semantic aspects within class, facilitating the division of known classes into sub-parts?
(3) How to obtain decision boundaries for these sub-parts within each known intent class?

To solve these challenges, we propose a \textbf{M}ulti-granularity \textbf{O}pen intent classification method via adaptive \textbf{G}ranular \textbf{B}all decision boundary (MOGB). Our MOGB method consists of a hierarchical representation learning module and a decision boundary acquiring module.
For \textbf{challenge (1)}, inspired by granular-ball computing~\cite{xia2023gbrs}, we propose a hierarchical representation approach through adaptive granular-ball clustering, which reflects intent distribution using multiple granular-balls with varying sizes.
For \textbf{challenge (2)}, the nearest sub-centroid classifier is proposed to learn discriminative representations, with the loss function encouraging each intent to move closer to its semantically closest sub-centroid.
In the training of hierarchical representation learning, the adaptive granular-ball clustering and nearest sub-centroid classification are iteratively conducted to mutually complement each other, facilitating the acquisition of discriminative representations.
For \textbf{challenge (3)}, we propose multi-granularity decision boundaries through granular-balls with varying radii and sub-centroids to distinguish the known and open intents. 
Our contributions are summarized as follows:
\begin{itemize}
\item We propose the hierarchical representation learning method, which adaptively reflects intra-class structure and intent distribution via granular-ball clustering.

\item We propose the nearest sub-centroid classifier to acquire discriminative representations by gathering intents to its semantically nearest sub-centroid.

\item We design multi-granularity decision boundaries for each known intent class to facilitate open intent classification, which provides a new way to solve this question.
\end{itemize}

\section{Related work}\label{sec:2}

\subsection{Open Intent Classification}
Recognizing the intent behind user interactions is crucial in dialogue systems, as they aim to understand users' potential requirements. 
Most classification models currently work under the closed-world assumption, which may not align with real-world systems functioning in open settings. Such systems often face queries that fall outside the supported intents, referred to as out-of-domain queries. 
With the prevalent of dialogue systems, there has been an increasing emphasis on identifying open intents in recent times~\cite{zhou2021contrastive,zhan2021out,parmar2023open,yan2020unknown,zhang2021discovering}.

Existing open intent classification methods can be categorized into two primary groups. The first group involves directly training a classifier with $K$+1 classes, with an extra class representing the open intent~\cite{zheng2020out}. 
The second group of methodologies focused on outlier detection algorithms, which can be categorized into two main subtypes: probability-based and decision boundary-based approaches. 
Probability-based methods use a predefined threshold to determine if a sample represents an open intent~\cite{shu2017doc}.
Boundary-based approaches~\cite{zhang2021deep,liu2023effective,zhang2023learning} have been developed to overcome the limitations of probability-based methods, which struggle to establish a distinct boundary between known classes and unknowns.

However, current boundary-based approaches assume that the feature distribution of known intents is within a compact spherical region, but this is not always the case in reality. In addition, the single decision boundary they established includes intra-open intents, increasing the open space risks.

\subsection{Granular-ball Computing}
Building on the human cognitive mechanisms of ``macro first''~\cite{chen1982topological} and multi-granularity cognitive computation~\cite{wang2017dgcc},
Xia et al.~\cite{xia2019granular} proposed granular-ball computing, which is efficient, robust, and interpretable. 
It adheres to a hierarchical rule of progressing from coarse-grained to fine-grained. In granular-ball computing, varying sizes of granular-balls are adaptively generated to depict data distribution. Each granular-ball is defined by a centroid and a radius, with the centroid representing the mean of all object vectors within the ball and the radius determined by the average distance of objects from the centroid. The effectiveness of this self-supervised data-covering method has been empirically validated~\cite{xie2024gbg++}, and it has demonstrated successful applications in addressing diverse challenges within the field of AI~\cite{liuunlock,wang2024text,wang2024gbrain}.

Current granular-ball computing methods encompass granular-ball classification~\cite{xia2024gbsvm,quadir2024granular}, granular-ball clustering~\cite{xie2024w,xie2024mgnr}, granular-ball three-way decision~\cite{yang20243wc,xia2024three}, and granular-ball rough sets~\cite{xia2020gbnrs,xia2023gbrs,zhang2023incremental}.
Furthermore, several applications have demonstrated the effectiveness of granular-ball representation, such as granular-ball representation in text adversarial defense~\cite{wang2024text}, label noise combating~\cite{wang2024gbrain}, feature selection~\cite{cao2024open}, and deep reinforcement learning~\cite{liuunlock}.

In summary, granular-ball computing is an effective way to discover the structure of arbitrary data distribution, with the radius of the granular-ball serving as a natural decision boundary for open classification. Despite this, there remains a notable absence of research applying granular-ball computing to the specific domain of open intent classification. 
Our MOGB method will employ granular-ball clustering to achieve hierarchical representation and establish multi-granularity adaptive decision boundaries.

\begin{figure*}[t]
\centering
\includegraphics[width=1.9\columnwidth]{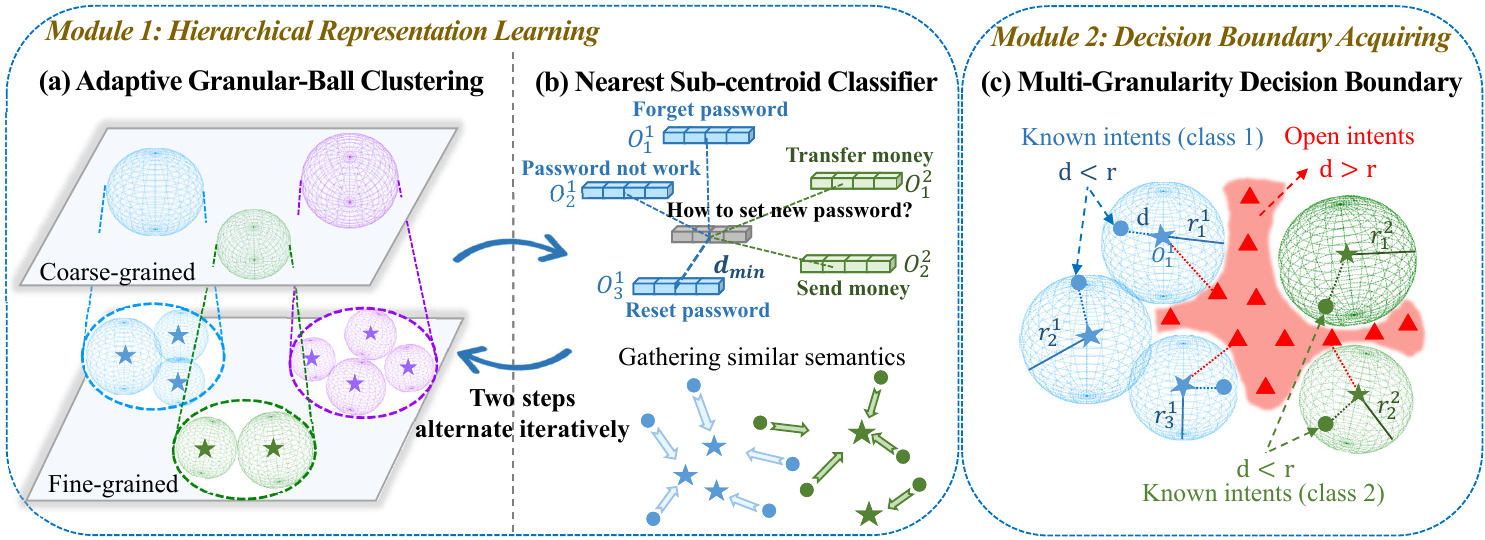}
\caption{The architecture of MOGB. In the hierarchical representation learning module, we generate granular-balls on all known intents via adaptive granular-ball clustering and then use the nearest sub-centroid classifier to learn representation. In the boundary acquiring module, multi-granularity decision boundaries are established.}
\label{framework}
\end{figure*}
\section{Our Methodology}\label{sec:4}
\subsection{Problem Statement and Model Overview}

\subsubsection{Problem Statement}
Given a dataset $S=\{(\mu_i, y_i)\}_{i=1}^{N}$ consisting of $N$ intents, where each pair $(\mu_i, y_i)$ represents an intent $\mu_i$ and its corresponding label $y_i$. The labels $y_i \in \{1,\cdots, K, K+1\}$ can be divided into two categories: $I_{\text{Known}}=\{1,\cdots, K\}$ denotes a set of known intent labels with $K$ representing the number of known intent classes, while $I_{\text{Unknown}}=K+1$ is used to label unknown (open) intents. During training and validation, only samples associated with known intents are utilized. In contrast, the testing dataset includes both known and open intents, and we need to classify the known intents into their respective classes and identify open intents as unknowns.
\subsubsection{Model Overview}
Our proposed MOGB method consists of two modules: hierarchical representation learning and decision boundary acquiring. In module 1, the hierarchical representation of each known intent class is obtained via adaptive granular-ball clustering, as depicted in Figure~\ref{framework} (a). Subsequently, we learn discriminative representation by the nearest sub-centroid classifier, as shown in Figure~\ref{framework} (b).
The iterative process of granular-ball clustering and nearest sub-centroid classification during training mutually reinforce each other to obtain discriminative representations. 
In module 2, multi-granularity decision boundaries are constructed for open intent classification, as illustrated in Figure~\ref{framework} (c).

\subsection{Hierarchical Representation Learning}
In this section, the BERT pre-trained model is employed for feature extraction, followed by adaptive granular-ball clustering and the nearest sub-centroid classifier.

\subsubsection{Feature Extraction}\label{sec:3.2.1}
Utilizing the pre-trained BERT model~\cite{kenton2019bert}, we extract token embeddings $[CLS, T_1, T_2, \ldots, T_M] \in \mathbb{R}^{(M+1) \times H}$ from the final hidden layer to represent each intent $\mu_i$. Here, $M$ represents the sequence length and $H$ denotes the embedding dimension. Following previous studies~\cite{zhang2021deep}, a mean pooling operation is applied to these embeddings in order to construct a semantic representation of each sentence denoted as $x_i \in \mathbb{R}^H$:
\begin{equation}
x_i = \text{MeanPooling}([CLS, T_1, T_2, \ldots, T_M]),
\end{equation}
where $CLS$ acts as a classification token. To further enhance the feature extraction process, $x_i$ is subsequently passed through a dense layer $h$, transforming the representation to $z_i \in \mathbb{R}^D$:
\begin{equation}
z_i = h(x_i) = \text{ReLU}(W_h x_i + b_h),
\end{equation}
where $D$ denotes the dimension of the intent representation, $W_h \in \mathbb{R}^{H \times D}$ and $b_h \in \mathbb{R}^D$ represent the weight and bias parameters of the dense layer $h$, respectively. 

\subsubsection{Adaptive Granular-ball Clustering}\label{sec:cluster}
In order to accurately represent the true distribution of the known intent class and mitigate open space risk associated with intra-open intents, we need to group the known intent class into sub-classes that reflect inherent patterns within the class.
Although conventional class-wise clustering methods can partially reveal data structure, their predefined cluster number setting limits the exploration of diverse patterns within each class. 
Granular-ball clustering is an efficient and adaptive clustering method to represent the true data distribution~\cite{xia2021granular}.
Motivated by it, we represent the intent distribution using granular-balls generated through adaptive granular-ball clustering, as shown in Figure~\ref{framework} (a), which eliminates the necessity for specifying cluster numbers. 

The basic definitions of the granular-ball are introduced as following~\cite{xia2021granular}:
\begin{definition}
Given a dataset of intents $Z=\{(z_i, y_i) \ |\  i=1, 2, \ldots,N\}$ consisting of $N$ feature representations $z_i$ and their corresponding labels $y_i$, we have a set of granular-balls denoted as $G = \{gb_1, gb_2, \ldots,  gb_m\}$ generated on the dataset, where $m$ denotes the number of granular-balls.
Each granular-ball $gb_j = \{(z_i, y_i)\ | \ i = 1, 2, \ldots, n_j\}$ represents a distinct subset containing $n_j$ intents. 
The fundamental components of each granular-ball include sample count $n_j$, centroid $O_j$, radius $r_j$, label $l_j$, and purity $p_j$.
The centroid of $gb_j$ is computed as the average representation $O_j = \frac{1}{n_j} \sum_{i=1}^{n_j} z_i$. 
The radius of $gb_j$ is determined by calculating the average distance between all samples and the centroid: $r_j = \frac{1}{n_j} \sum_{i=1}^{n_j} ||z_i - O_j||$, where $||z_i - O_j||$ denotes the Euclidean distance between $z_i$ and $O_j$.
The label $l_j$ assigned to granular-ball $gb_j$ corresponds to the class that contains the highest number of samples within it.
Purity $p_j$ indicates the proportion of samples labeled with $l_j$ in granular-ball $gb_j$.
\end{definition}

The input of the clustering includes all representations $Z = \{(z_i, y_i) \ |\  i = 1, 2, \ldots, N\}$ from different known intent classes. 
Adaptive granular-ball clustering is an iterative process to split granular-balls. 
First, the representations $Z$ are initialized as a single granular-ball. Subsequently, the granular-balls that satisfy the split conditions undergo splitting (conditions described below). The process terminates when all granular-balls can no longer be further split.

To control the splitting process effectively, we set the condition of purity limit $p_l$ and sample count limit $n_l$.
Specifically, for a granular-ball $gb_j$ with purity $p_j$ and sample count $n_j$, if $p_j<p_l$ and $n_j>n_l$, the granular-ball will be split further until either $p_j \ge p_l$ or $n_j \le n_l$. 
During the splitting phase, the granular-ball with a unique class label set $L_j$ will be split into $|L_j|$ new granular-balls, where $L_j=\text{unique}(\{y_i | (z_i,y_i) \in gb_j\})$. 
To achieve this division accurately, one intent from each unique class within the original granular-ball is randomly selected as a pseudo centroid for creating a new granular-ball.
Subsequently, other intents are assigned to their closest respective new granular-balls based on their Euclidean distances from pseudo centroids.

The adaptive granular-ball clustering algorithm produces a set of granular-balls denoted as $G = \{gb_1, gb_2, \ldots, gb_m\}$, where each granular-ball $gb_j$ ($j=1,\cdots,m$) is assigned a label $l_j \in I_{\text{Known}}=\{1,\cdots, K\}$. 
To eliminate outliers and prevent overfitting, we select high-quality granular-balls with purity $p_j$ above $p_t$ and sample count $n_j$ more than $n_t$ to represent known intents.
Each known intent class $c \in I_{\text{Known}}$ corresponds to a number of $n_c$ granular-balls labeled with $c$, where $n_c=count(l_j=c)$, and $\sum_{c=1}^{K} n_c =m$. The granular-balls representing class $c$ are denoted as $\{gb_{s}^{c} \}_{s=1}^{n_c}$, where $gb_{s}^{c}$ represents the $s$-th granular-ball with label $c$. The collection of all these granular-balls $\{gb_{s}^{c}\}_{c,s=1}^{K,n_c}$ reveals the detailed structure of the distributions for all known intents. 

\subsubsection{Nearest Sub-Centroid Classifier}
The effectiveness of adaptive granular-ball clustering is highly dependent on the quality of representation. Therefore, it is necessary to obtain discriminative representations that reflect the semantic structure of known intent classes.
However, previous cross-entropy loss-based~\cite{zhang2021deep} or contrastive loss-based~\cite{liu2023effective} representation learning methods treat each class as an entity, failing to explore patterns within the class.
To this end, we propose the nearest sub-centroid representation learning method to gather similar semantics and separate dissimilar semantics, as shown in Figure~\ref{framework} (b).

According to the granular-balls $\{gb_{s}^{c}\}_{c,s=1}^{K,n_c}$, we compute their centroids as $\{O_{s}^{c} \}_{c,s=1}^{K,n_c}$, where $O_{s}^{c}$ denotes the $s$-th sub-centroids of class $c$. 
These sub-centroids serve as the informative properties of known intents.
In the classification of representation learning, the distances from test intent to all sub-centroids are calculated as follows:
\begin{equation}
   dis\left\langle z_i, O_{s}^{c} \right\rangle = \frac{ z_i^\top O_{s}^{c}}{\| z_i\| \|O_{s}^{c}\|}, \ \ i=\{1,\cdots,N\},
\end{equation}
where $\|z_i\|$ and $\|O_{s}^{c}\|$ denote the Euclidean norm vectors of $z_i$ and $O_{s}^{c}$. We assign the label of test intent $z_i$ corresponding to the label of the sub-centroid closest to $z_i$. 
Concretely, if $ dis\left\langle z_i, O_{s^*}^{c^*} \right\rangle$ denotes the minimum distance among $ dis\left\langle z_i, O_{s}^{c} \right\rangle$, the predicted label of $z_{i}$ is $c^*$. The classifier rule is elaborated as follows:
\begin{equation}
\hat{y}_i = c^*,(c^*, s^*) = \underset{ c \in \{1, \dots, K\}, s \in \{1, \dots, n_c\} }{\arg\min}  dis\left\langle z_i, O_{s}^{c} \right\rangle.
\end{equation}

Based on the classifier provided, the training loss associated with the nearest sub-centroid representation learning is shown as follows:
\begin{equation}\label{loss:cluster}
\mathcal{L}_{gb} = \frac{1}{N} \sum_{i=1}^{N} -\log p(y_{i} | z_{i}),
\end{equation}

\begin{equation}
\quad p(y | z) = \frac{\exp \left(-\min \left( \left \{ dis\left \langle z,O_{s}^{y}   \right \rangle  \right \} _{s=1}^{n_y} \right) \right)}{\sum_{c=1}^{K}  \exp \left( - \min \left( \left \{ dis\left \langle z,O_{s}^{c}   \right \rangle  \right \} _{s=1}^{n_c} \right) \right)}.
\end{equation}

During hierarchical representation learning, adaptive granular-ball clustering produces informative sub-centroids and the nearest sub-centroid classifier optimizes the representation by adjusting the arrangement between intents and sub-centroids.
The improved representation subsequently facilitates the identification of more informative sub-centroids. In summary, the two steps iteratively complement each other, eventually benefiting the open intent classification.

\subsection{Multi-Granularity Decision Boundary in Inference}
In our MOGB method, pairs of centroids and radii $\{O_{s}^{c}, r_{s}^{c}\}_{c,s=1}^{K,n_c}$ generated from training data of known intents are directly employed to construct multi-granularity decision boundaries for each class. 

In inference, the distances between test sample $z_i$ and all sub-centroids $\{O_{s}^{c}\}_{c,s=1}^{K,n_c}$ are first calculated as $dis \langle z_i, O_s^c \rangle$.
Then, we compare each $dis \langle z_i, O_s^c \rangle$ with its corresponding radii $\{r_{s}^{c}\}_{c,s=1}^{K,n_c}$ to make open classificatioin.
If the test sample falls outside all multi-granularity boundaries, we classify it as unknown. Otherwise, the sample is predicted as the same class as the nearest sub-centroid that satisfies the distance condition. The inference rule is formulated as:
\begin{equation}
\hat{y_i} = 
\begin{cases}
\text{unknown}, \text{if }  \ dis  \langle z_i, O_s^c \rangle > r_s^c, \ \forall c \in I_{\text{Known}}; \\
\underset{c \in I_{\text{Known}},dis \langle z_i, O_s^c \rangle \leq r_s^c }{\arg\min} \, dis \langle z_i, O_s^c \rangle, \text{Otherwise}.
\end{cases}
\end{equation}

Intuitively, compared to existing boundary-based methods that include two training processes, our MOGB method can simultaneously learn representation and acquire decision boundaries with just one training.

\section{Experiments}\label{sec:5}
\subsection{Datasets}
In accordance with prior studies~\cite{zhang2021deep,zhang2023learning}, we conduct experiments on three commonly used datasets: 
 \textbf{StackOverﬂow}~\cite{xu2015short} is a dataset containing 3,370,528 programming question titles across 20 categories. Consistent with prior studies, a subset of 1,000 examples from each category was randomly selected for analysis. \textbf{SNIPS}~\cite{coucke2018snips} is composed of 14,484 spoken utterances with 7 distinct categories of intent classes. 
\textbf{BANKING}~\cite{casanueva2020efficient} is an online banking inquiry with 13,080 instances and 77 intents classes.
 Table~\ref{tab:1} presents detailed statistics of these datasets.

\subsection{Baselines}
We compare our MOGB method with the following open intent classification methods:
\textbf{MSP}~\cite{hendrycks2022baseline} classifies samples based on their maximum softmax probabilities; 
\textbf{DOC}~\cite{shu2017doc} conducts open intent classification by establishing probability thresholds through Gaussian fitting. 
\textbf{OpenMax}~\cite{bendale2016towards} implements open set recognition by transforming the softmax layer in a neural network into an OpenMax layer. 
\textbf{DeepUnk}~\cite{lin2019deep} utilizes margin loss that minimizes intra-class variance and maximizes inter-class variance to learn enhanced representation.
\textbf{ADB}~\cite{zhang2021deep}  acquires feature representation by softmax loss and learns a single decision boundary for each class.
\textbf{DA-ADB}~\cite{zhang2023learning} learns distance-aware intent representations to obtain appropriate decision boundary.
\subsection{Evaluation Metrics}
Following previous work~\cite{zhang2021deep}, we utilize F1-score and accuracy metrics to assess different approaches. Additionally, we also compute the fine-grained macro F1-score for both known classes and unknowns.
\begin{table}[t]
\centering
 
\begin{tabular}{lccc}

\hline
Dataset &\#Class&\#Train/Valid/Test&Length\\
\hline
StackOverflow & 20 &12,000/2,000/6,000&9.18 \\
SNIPS         & 7  &13,084/700/700&9.05 \\
BANKING       & 77 &9,003/1,000/3,080&11.91 \\
\hline
\end{tabular}
\caption{Statistics of datasets.}
\label{tab:1}
\end{table}
\begin{table*}[t]
\centering
\begin{tabular}{lc|cccc|cccc|cccc}
\hline
& & \multicolumn{4}{c }{StackOverflow} & \multicolumn{4}{c }{SNIPS} & \multicolumn{4}{c }{BANKING} \\ \cmidrule(r){3-6} \cmidrule(r){7-10} \cmidrule(r){11-14}   
 &Methods         & Acc   & F1-All    & F1-U     & F1-K         & Acc    & F1-All     & F1-U  & F1-K       & Acc      & F1-All      & F1-U     & F1-K  \\ \hline
\multirow{7}{*}{25\%} 
&MSP$\dagger$     &28.67  & 37.85  & 13.03   & 42.82    &28.57  &37.65  &0.00    &56.48     & 43.67    & 50.09   & 41.43      & 50.55  \\
&DOC$\dagger$     &42.74  & 47.73  & 41.25   & 49.02    &45.63  &55.10  &34.67   &65.32     & 56.99    & 58.03   & 61.42      & 57.85  \\
&OpenMax$\dagger$ &40.28  & 45.98  & 49.29   & 52.60    &59.57  &49.68  &61.92   &43.56     & 49.94    & 54.14   & 51.32      & 54.23  \\
& DeepUnk$\ast$   &70.68  & 65.57  & 36.87   & 47.39    &53.42  &59.67  &49.32   &64.85     & 70.68    & 65.57   & 76.98       &64.97\\

&ADB$\dagger$     &86.72  & 80.83  & 90.88   & 78.82    &58.57  &65.72  &58.92 &69.12& 78.85  & 71.62   & 84.56  & 70.94  \\
& DA-ADB$\ast$    &87.01  & 80.05  & 91.16   &77.82     &47.09  &52.75  &44.38   &56.93     & 78.39    & 70.67   & 84.27         &69.95\\ 
&\textbf{MOGB}    &\textbf{91.48} &\textbf{84.43}&\textbf{94.42} &\textbf{87.80}&\textbf{68.00} &\textbf{72.20}&\textbf{71.36}  &\textbf{72.63}& \textbf{83.08}& \textbf{75.19}&\textbf{88.29}& \textbf{74.50}\\
                      \hline

\multirow{7}{*}{50\%} 
&MSP$\dagger$& 52.42    & 63.01  & 23.99   & 66.91&57.97      &63.34   &7.12      &77.40   & 59.73    &  63.60&  41.19&  71.97\\
&DOC$\dagger$& 52.53    & 62.84    & 25.44   & 66.58&72.50      &78.15  &54.89     &\textbf{83.97}     & 64.81    & 73.12 & 55.14  & 73.59\\
&OpenMax$\dagger$& 60.35    & 68.18        & 45.00   & 70.49&59.82      &65.50   &14.23     &78.31  & 65.31    & 74.24  & 54.33  & 74.76    \\
& DeepUnk$\ast$ & 71.01& 75.41& 35.80& 67.67& 69.53& 75.10& 49.67& 81.45& 71.01& 75.41& 67.80&75.61\\

&ADB$\dagger$& 86.40    & 85.83  & 87.34   & 85.68&73.57      &78.82    &62.17     &82.98       & 78.86    & 80.90     & 78.44  & 80.96   \\
& DA-ADB$\ast$ & 86.02& 85.21& 87.21& 85.01& 73.69& 77.35& \textbf{66.58}& 80.05& 79.09& 81.03& 78.78&81.09\\
&\textbf{MOGB} &\textbf{88.67} &\textbf{87.49} &\textbf{89.71} &\textbf{87.27}&\textbf{73.86}&\textbf{78.91} & \underline{\textbf{62.42}}&\underline{\textbf{83.03}}  & \textbf{80.58}& \textbf{81.52}&\textbf{81.04} &\textbf{81.53}\\
                      \hline

\multirow{7}{*}{75\%} &MSP$\dagger$& 72.17    & 77.95   & 33.96   & 80.88   &71.49     &72.56     &6.52     &85.77      & 75.89    & 83.60  & 39.23  & 84.36  \\
&DOC$\dagger$& 68.91    & 75.06   & 16.76   & 78.95&79.96     &83.11      & 51.67    &89.40   & 76.77    & 83.34  & 50.60  & 83.91   \\
&OpenMax$\dagger$& 
74.42    & 79.78    & 44.87   & 82.11&72.20     &73.63   &11.03     &86.15         & 77.45    & 84.07  & 50.85  & 84.64 \\
& DeepUnk$\ast$ & 71.56& 77.63& 34.38& 77.63 & 78.50& 81.74& 52.05& 87.68& 74.73& 81.12& 50.57&81.65\\

 &ADB$\dagger$ & 82.78    & 85.99    & 73.86   & 86.80&86.71     &88.61   &73.35      &91.66        & 81.08& 85.96&66.47& 86.09\\
& DA-ADB$\ast$ & 82.89& 86.11& 74.06& 86.92& 81.57& 84.29& 68.62& 87.42& 81.31& 86.01& 67.22&86.36\\ 
&\textbf{MOGB}   &\textbf{84.37} &\textbf{87.37}&\textbf{75.52} &\textbf{88.16}&\textbf{87.71} & \textbf{89.39} &\textbf{73.62} &\textbf{92.55}& \textbf{82.69} & \textbf{86.82}  & \textbf{71.27}  &\textbf{87.09} \\
                       \hline
\end{tabular}
\caption{Results of MOGB across StackOverflow, SNIPS, and BANKING with different known class ratios (25\%, 50\%, and 75\%). $\ast$ means the results are not provided in the original paper and we get the results by running its released codes.
The results with $\dagger$ are from~\cite{zhang2021deep}.}
\label{tab:2}
\end{table*}
\subsection{Experimental Settings}
Following the experimental setting of prior study~\cite{zhang2021deep}, a predetermined percentage of classes are randomly designated as known classes, while the remaining classes are considered unknown (open). During testing, all unknown classes are collectively treated as a single additional class. Training exclusively utilizes samples from the known classes, with samples from the unknown classes excluded from the training dataset. The test dataset comprises both known and unknown samples. The experimental results are presented based on three datasets, with known class ratios set at 25\%, 50\%, and 75\%, respectively.

During representation learning, we use the pre-trained BERT model \cite{kenton2019bert}, keeping all parameters fixed except for those in the final layer. The training batch size is set to 128, with a learning rate of 2e-5 to fine-tune the final layer. We control the adaptive granular-ball clustering by two attributes of the granular-ball: purity limit $p_l$ and sample count limit $n_l$. We set $p_l=0.9$ for all datasets and varying $n_l$ for different datasets depending on the number of samples within each class. In addition, granular-balls with high quality are selected to represent the distribution by new purity limit $p_t$ and sample count limit $n_t$. We set $p_t=1$ for all datasets and set varying $n_l$ based on the dataset.

\subsection{Main Results}

Table~\ref{tab:2} displays the results of open intent classification on three datasets (StackOverflow, SNIPS, and BANKING) at different proportions of known classes (25\%, 50\%, and 75\%). 
We report the accuracy of all intent classes (denoted as ``Acc'') and F1-score for all intent classes, unknown, and known classes (denoted as ``F1-All'', ``F1-U'', and ``F1-K'', respectively).
The most optimal performance is emphasized through the \textbf{blod} values, while the \underline{\textbf{underline}} values indicate a suboptimal performance of our MOGB.

According to Acc and F1-All, our MOGB methodology exhibits superior performance across all three datasets, consistently surpassing all other baseline methods. 
Moreover, we observed that our MOGB approach surpasses ADB and DA-ADB by a greater margin when the known class ratio is low. Conversely, when the known class ratio is high, the enhancement in performance of our method is somewhat minimal. 
Specifically, in cases where the proportion of the known class is low (25\%), compared with the strong baseline ADB, our MOGB improves four metrics (``Acc'' / ``F1-All'' / ``F1-U'' / ``F1-K'') by 4.76\% / 3.60\% / 3.54\% / 8.98\% on StackOverflow, by 9.43\% / 6.48\% / 12.44\% / 3.51\% on SNIPS, and by 4.23\% / 3.57\% / 3.73\% / 3.56\% on BANKING. Conversely, when the known class ratio is higher (75\%), compared with ADB, our MOGB method enhances four metrics by 1.59\% / 1.38\% / 1.66\% / 1.36\% on StackOverflow, by 1.00\% / 0.78\% / 0.27\% / 0.89\% on SNIPS and by 1.61\% / 0.86\% / 4.80\% / 1.00\% on BANKING. 

These above results indicate that our MOGB is more effective in situations with a high proportion of unknown intents. 
This can be ascribed to the fact that the test dataset includes more open intents under a low known class ratio. Compared with the single decision boundary methods (ADB and DA-ADB), the multi-granularity decision boundaries in MOGB allocate more space for open intents. Therefore, the open space risk associated with intra-open intents is reduced effectively, resulting in enhanced performance.

\subsection{Ablation Study}
In this subsection, we examine the impact of two components of MOGB on three datasets with a known class ratio of 25\%. From Table~\ref{tab:3}, we observed that removing any component will lead to performance degradation, emphasizing the essence of each independent component. Specifically, (1) w/o HRL refers to removing the Hierarchical Representation Learning (HRL) module and learning representation by cross-entropy loss; (2) w/o MB refers to removing the Multi-granularity decision Boundary (MB) and classification by a single boundary of each class; (3) w/o HRL+MB indicates the exclusion of both Hierarchical Representation Learning and Multi-granularity decision Boundary. The combination of HRL and MB can achieve the best performance.

\begin{table}[t]
\centering
\begin{tabular}{l l c c}
\toprule
Dataset & Method       & Acc & F1-All \\
\midrule
\multirow{4}{*}{StackOverflow} 
    & MOGB        & \textbf{91.48}  & \textbf{84.43} \\
    & w/o HRL      & 85.70           & 79.72          \\
    & w/o MB       & 82.38           & 75.82          \\
    & w/o HRL+MB   & 86.30           & 80.06          \\
\midrule
\multirow{4}{*}{SNIPS} 
    & MOGB              & \textbf{68.00}         & \textbf{72.20} \\
    & w/o HRL      & 62.57           & 68.94          \\
    & w/o MB       & 62.57           & 68.88          \\
    & w/o HRL+MB   & 60.29           & 67.54          \\
\midrule
\multirow{4}{*}{BANKING} 
    & MOGB              & \textbf{83.08}  & \textbf{75.19} \\
    & w/o HRL      & 77.40           & 71.55          \\
    & w/o MB       & 81.14           & 74.25          \\
    & w/o HRL+MB   & 78.80           & 72.20          \\
\bottomrule
\end{tabular}
\caption{Ablation study on StackOverflow, SNIPS, and BANKING with a known class ratio of 25\%.}
\label{tab:3}
\end{table}

\section{Discussion}\label{sec:6}

In the MOGB method, the multi-granularity decision boundaries are established based on granular-balls derived from the training dataset. Consequently, the quality of these granular-balls has a significant impact on the classification performance. The quality of a granular-ball largely depends on its purity and the sample count. We use granular-balls with purity exceeding $p_t$ and sample count more than $n_t$ to build the decision boundaries. Next, we report the effect of $p_t$ and $n_t$.

\subsection{Effect of the Purity Limit $p_t$}
A higher $p_t$ results in a small number of trustworthy granular-balls being used to represent the dataset and fewer decision boundaries being constructed, reducing the known intent space and increasing the chance of misclassifying known intents as unknowns. Conversely, a lower $p_t$ expands the known space, reducing open space but increasing the risk of misclassifying unknown intents as known classes.
\begin{table}[t]
\centering
\begin{tabular}{c|cc cc }
\hline
$p_t$  & Acc & F1-All   & F1-U & F1-K\\ \hline
0.80 & 83.10 & 86.41 & 72.70 & 87.32 \\
0.85 & 83.27 & 86.50 & 73.15 & 87.39 \\
0.90 & 84.37 & 87.37 & 75.52 & 88.16 \\
0.93 & \textbf{84.45} & \textbf{87.41} & \textbf{75.78} & \textbf{88.19} \\
0.95 & 84.23 & 87.31 & 75.40 & 88.10 \\
0.97 & 84.23 & 87.31 & 75.40 & 88.10 \\
\hline
\end{tabular}
\caption{Effect of $p_t$ on StackOverflow with known class ratio of 75\%. }
\label{tab:4}
\end{table}

Table~\ref{tab:4} presents the Acc, F1-All, F1-U, and F1-K as the purity threshold \(p_t\) increases from 0.80 to 0.97 on the StackOverflow dataset, with a known class ratio of 75\%. Overall, the performance of our method in open classification remains relatively stable, highlighting its robustness. However, a slight trend in performance variation can be observed: as \(p_t\) increases, classification performance initially improves, peaking at \(p_t = 0.93\). This improvement occurs because high-purity granular-balls better capture the distribution of known classes by reducing uncertainty. Beyond this point, further increasing \(p_t\) to 0.95 causes a performance decline. The stricter purity requirement excludes certain decision boundaries that represent known classes, leading to portions of known class spaces being misclassified as unknowns and increasing open space risk. This outcome is consistent with our expectations.

\subsection{Effect of the Sample Count Limit $n_t$}
\begin{figure}[t]
\centering
\includegraphics[width=1.0\columnwidth]{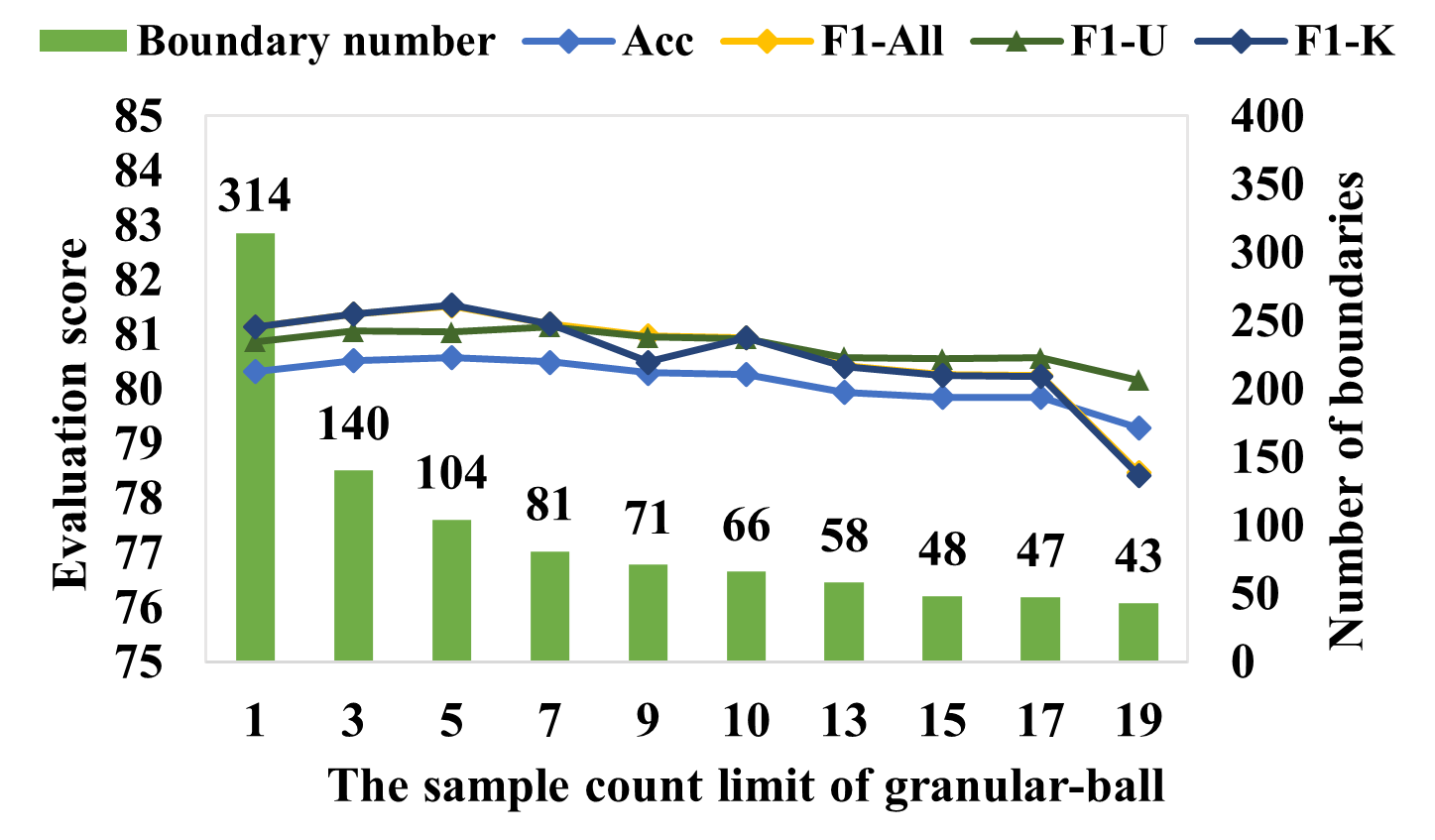} 
\caption{Effect of $n_t$ on BANKING with 50\% known class. X-axis represents the value of $n_t$, the left Y-axis denotes the values of four metrics, and the right Y-axis indicates the number of established decision boundaries for all known classes.}
\label{fig:3}
\end{figure}

Granular-balls containing samples exceeding a specified threshold $n_t$ are chosen to represent data distribution and establish decision boundaries for the open classification. 
Figure~\ref{fig:3} shows the performance of MOGB on dataset BANKING with a known class ratio of 50\% as $n_t$ changes, along with the number of established decision boundaries at different $n_t$ values. 
There are 39 known classes under a known class ratio of 50\% on the BANKING dataset. From Figure~\ref{fig:3}, we observed that the number of decision boundaries for open classification decreases as \(n_t\) increases. 
In particular, when \( n_t = 1 \), a total of 314 decision boundaries are established, signifying that each known intent class is characterized by several distinct boundaries. Optimal performance is achieved at \(n_t = 5\), where 104 decision boundaries are constructed. However, at \(n_t=19\), only 43 decision boundaries are constructed, with most classes represented by a single boundary, degrading the performance of four metrics. 

The observed phenomenon aligns with our expectations. A strict \(n_t\) limit reduces decision boundaries, leading to incomplete feature space coverage and higher empirical risk. In contrast, a looser \(n_t\) limit introduces more boundaries that better fit the true distribution but risk overcovering, ultimately reducing performance. Although the performance of MOGB varies with changes in \(n_t\), the overall fluctuation remains small, demonstrating the robustness of our method.

\section{Conclusions}\label{sec:7}

This paper proposes a multi-granularity open intent classification method via hierarchical representation learning and multi-granularity decision boundary (MOGB). Specifically, the granular-balls of diverse sizes are generated by adaptive granular-ball clustering to represent the known intent space during hierarchical representation learning. Additionally, the nearest sub-centroid classifier equips for learning fine-grained representation that reflects semantic structures within known intent classes. Furthermore, multi-granularity decision boundaries for each known intent class are constructed for open intent classification, which reduces both empirical and open-space risk. Finally, extensive experiments demonstrate the superiority of the MOGB method.

\section*{Acknowledgments}
This work was supported by the National Natural Science Foundation of China (Nos. 62476228, 62406259, 62222601, 62450043, 62176033), the Sichuan Science and Technology Program (Nos. 2024ZYD0180, 2024YFHZ0024, MZGC20240153), and the Graduate Representative Achievement Cultivation Project of Southwest University of Finance and Economics (Nos. JGS2024066, JGS2024068).
\bibliography{aaai25}

\begin{thebibliography}{40}
\providecommand{\natexlab}[1]{#1}

\bibitem[{Bendale and Boult(2016)}]{bendale2016towards}
Bendale, A.; and Boult, T.~E. 2016.
\newblock Towards open set deep networks.
\newblock In \emph{Proceedings of the IEEE Conference on Computer Vision and Pattern Recognition}, 1563--1572.

\bibitem[{Cao et~al.(2024)Cao, Yang, Xia, Wang, and Li}]{cao2024open}
Cao, X.; Yang, X.; Xia, S.; Wang, G.; and Li, T. 2024.
\newblock Open Continual Feature Selection via Granular-Ball Knowledge Transfer.
\newblock \emph{IEEE Transactions on Knowledge and Data Engineering}, 36(12): 8967--8980.

\bibitem[{Casanueva et~al.(2020)Casanueva, Tem{\v{c}}inas, Gerz, Henderson, and Vuli{\'c}}]{casanueva2020efficient}
Casanueva, I.; Tem{\v{c}}inas, T.; Gerz, D.; Henderson, M.; and Vuli{\'c}, I. 2020.
\newblock Efficient intent detection with dual sentence encoders.
\newblock In \emph{Proceedings of the Workshop on Natural Language Processing for Conversational AI}, 38--45.

\bibitem[{Chen(1982)}]{chen1982topological}
Chen, L. 1982.
\newblock Topological structure in visual perception.
\newblock \emph{Science}, 218(4573): 699--700.

\bibitem[{Cheng et~al.(2022)Cheng, Jiang, Yin, Wang, and Gu}]{cheng2022learning}
Cheng, Z.; Jiang, Z.; Yin, Y.; Wang, C.; and Gu, Q. 2022.
\newblock Learning to classify open intent via soft labeling and manifold mixup.
\newblock \emph{IEEE/ACM Transactions on Audio, Speech, and Language Processing}, 30: 635--645.

\bibitem[{Coucke et~al.(2018)Coucke, Saade, Ball, Bluche, Caulier, Leroy, Doumouro, Gisselbrecht, Caltagirone, Lavril et~al.}]{coucke2018snips}
Coucke, A.; Saade, A.; Ball, A.; Bluche, T.; Caulier, A.; Leroy, D.; Doumouro, C.; Gisselbrecht, T.; Caltagirone, F.; Lavril, T.; et~al. 2018.
\newblock Snips voice platform: an embedded spoken language understanding system for private-by-design voice interfaces.
\newblock arXiv:1805.10190.

\bibitem[{Guo et~al.(2017)Guo, Pleiss, Sun, and Weinberger}]{guo2017calibration}
Guo, C.; Pleiss, G.; Sun, Y.; and Weinberger, K.~Q. 2017.
\newblock On calibration of modern neural networks.
\newblock In \emph{International Conference on Machine Learning}, 1321--1330.

\bibitem[{Hendrycks and Gimpel(2022)}]{hendrycks2022baseline}
Hendrycks, D.; and Gimpel, K. 2022.
\newblock A Baseline for Detecting Misclassified and Out-of-Distribution Examples in Neural Networks.
\newblock In \emph{International Conference on Learning Representations}.

\bibitem[{Jain, Scheirer, and Boult(2014)}]{jain2014multi}
Jain, L.~P.; Scheirer, W.~J.; and Boult, T.~E. 2014.
\newblock Multi-class open set recognition using probability of inclusion.
\newblock In \emph{Computer Vision--ECCV}, 393--409. Springer.

\bibitem[{Kenton and Toutanova(2019)}]{kenton2019bert}
Kenton, J. D. M. W.~C.; and Toutanova, L.~K. 2019.
\newblock {Bert: Pre-training of deep bidirectional transformers for language understanding}.
\newblock In \emph{Proceedings of NAACL-HLT}, volume~1, 2.

\bibitem[{Lin and Xu(2019)}]{lin2019deep}
Lin, T.~E.; and Xu, H. 2019.
\newblock Deep unknown Intent Detection with Margin Loss.
\newblock In \emph{Proceedings of the Annual Meeting of the Association for Computational Linguistics}, 5491--5496.

\bibitem[{Liu et~al.(2024)Liu, Jianye, Ma, and Xia}]{liuunlock}
Liu, J.; Jianye, H.; Ma, Y.; and Xia, S. 2024.
\newblock Unlock the Cognitive Generalization of Deep Reinforcement Learning via Granular Ball Representation.
\newblock In \emph{International Conference on Machine Learning}.

\bibitem[{Liu et~al.(2023)Liu, Li, Mu, Yang, Xu, and Wang}]{liu2023effective}
Liu, X.~K.; Li, J.~Q.; Mu, J.~J.; Yang, M.; Xu, R.~F.; and Wang, B.~Y. 2023.
\newblock Effective open intent classification with {K-center} contrastive learning and adjustable decision boundary.
\newblock In \emph{Proceedings of the AAAI Conference on Artificial Intelligence}, 13291--13299.

\bibitem[{Parmar et~al.(2023)Parmar, Chouhan, Raychoudhury, and Rathore}]{parmar2023open}
Parmar, J.; Chouhan, S.; Raychoudhury, V.; and Rathore, S. 2023.
\newblock Open-world machine learning: applications, challenges, and opportunities.
\newblock \emph{ACM Computing Surveys}, 55(10): 1--37.

\bibitem[{Quadir and Tanveer(2024)}]{quadir2024granular}
Quadir, A.; and Tanveer, M. 2024.
\newblock Granular ball twin support vector machine with pinball loss function.
\newblock \emph{IEEE Transactions on Computational Social Systems}, 1--10.

\bibitem[{Shu, Xu, and Liu(2017)}]{shu2017doc}
Shu, L.; Xu, H.; and Liu, B. 2017.
\newblock {DOC}: Deep open classification of text documents.
\newblock In \emph{Proceedings of th Conference on Empirical Methods in Natural Language Processing}, 2911--2916.

\bibitem[{Wang(2017)}]{wang2017dgcc}
Wang, G.~Y. 2017.
\newblock {DGCC}: data-driven granular cognitive computing.
\newblock \emph{Granular Computing}, 2(4): 343--355.

\bibitem[{Wang et~al.(2024{\natexlab{a}})Wang, Li, Xia, Lin, and Wang}]{wang2024text}
Wang, Z.; Li, J.; Xia, S.; Lin, L.; and Wang, G. 2024{\natexlab{a}}.
\newblock Text adversarial defense via granular-ball sample enhancement.
\newblock In \emph{Proceedings of the International Conference on Multimedia Retrieval}, 348--356.

\bibitem[{Wang et~al.(2024{\natexlab{b}})Wang, Zhang, Xia, Lin, and Wang}]{wang2024gbrain}
Wang, Z.~L.; Zhang, T.; Xia, S.~Y.; Lin, L.~L.; and Wang, G.~Y. 2024{\natexlab{b}}.
\newblock {GBRAIN}: Combating textual label noise by granular-ball based robust training.
\newblock In \emph{Proceedings of the International Conference on Multimedia Retrieval}, 357--365.

\bibitem[{Xia et~al.(2024{\natexlab{a}})Xia, Wang, Zhang, Yang, and Xia}]{xia2024three}
Xia, D.; Wang, G.; Zhang, Q.; Yang, J.; and Xia, S. 2024{\natexlab{a}}.
\newblock Three-way approximations fusion with granular-ball computing to guide multi-granularity fuzzy entropy for feature selection.
\newblock \emph{IEEE Transactions on Fuzzy Systems}, 32(10): 5963--5977.

\bibitem[{Xia et~al.(2024{\natexlab{b}})Xia, Lian, Wang, Gao, Chen, and Peng}]{xia2024gbsvm}
Xia, S.; Lian, X.; Wang, G.; Gao, X.; Chen, J.; and Peng, X. 2024{\natexlab{b}}.
\newblock Gbsvm: an efficient and robust support vector machine framework via granular-ball computing.
\newblock \emph{IEEE Transactions on Neural Networks and Learning Systems}, 1--15.

\bibitem[{Xia et~al.(2020)Xia, Zhang, Li, Wang, Giem, and Chen}]{xia2020gbnrs}
Xia, S.; Zhang, H.; Li, W.; Wang, G.; Giem, E.; and Chen, Z. 2020.
\newblock GBNRS: A novel rough set algorithm for fast adaptive attribute reduction in classification.
\newblock \emph{IEEE Transactions on Knowledge and Data Engineering}, 34(3): 1231--1242.

\bibitem[{Xia et~al.(2021)Xia, Zheng, Wang, Gao, and Wang}]{xia2021granular}
Xia, S.; Zheng, S.; Wang, G.; Gao, X.; and Wang, B. 2021.
\newblock Granular ball sampling for noisy label classification or imbalanced classification.
\newblock \emph{IEEE Transactions on Neural Networks and Learning Systems}, 34(4): 2144--2155.

\bibitem[{Xia et~al.(2019)Xia, Liu, Ding, Wang, Yu, and Luo}]{xia2019granular}
Xia, S.~Y.; Liu, Y.~S.; Ding, X.; Wang, G.~Y.; Yu, H.; and Luo, Y.~G. 2019.
\newblock Granular ball computing classifiers for efficient, scalable and robust learning.
\newblock \emph{Information Sciences}, 483: 136--152.

\bibitem[{Xia et~al.(2023)Xia, Wang, Wang, Gao, Ding, Yu, Zhai, and Chen}]{xia2023gbrs}
Xia, S.~Y.; Wang, C.; Wang, G.~Y.; Gao, X.~B.; Ding, W.~P.; Yu, J.~H.; Zhai, Y.~J.; and Chen, Z.~Z. 2023.
\newblock {GBRS}: A unified granular-ball learning model of pawlak rough set and neighborhood rough set.
\newblock \emph{IEEE Transactions on Neural Networks and Learning Systems}, 1--15.

\bibitem[{Xie et~al.(2024{\natexlab{a}})Xie, Hua, Xia, Cheng, Wang, and Gao}]{xie2024w}
Xie, J.; Hua, C.; Xia, S.; Cheng, Y.; Wang, G.; and Gao, X. 2024{\natexlab{a}}.
\newblock W-GBC: An Adaptive Weighted Clustering Method Based on Granular-Ball Structure.
\newblock In \emph{2024 IEEE 40th International Conference on Data Engineering (ICDE)}, 914--925.

\bibitem[{Xie et~al.(2024{\natexlab{b}})Xie, Xiang, Xia, Jiang, Wang, and Gao}]{xie2024mgnr}
Xie, J.; Xiang, X.; Xia, S.; Jiang, L.; Wang, G.; and Gao, X. 2024{\natexlab{b}}.
\newblock {MGNR}: A multi-granularity neighbor relationship and its application in {KNN} classification and clustering methods.
\newblock \emph{IEEE Transactions on Pattern Analysis and Machine Intelligence}, 46(12): 7956--7972.

\bibitem[{Xie et~al.(2024{\natexlab{c}})Xie, Zhang, Xia, Zhao, Wu, Wang, and Ding}]{xie2024gbg++}
Xie, Q.; Zhang, Q.; Xia, S.; Zhao, F.; Wu, C.; Wang, G.; and Ding, W. 2024{\natexlab{c}}.
\newblock Gbg++: A fast and stable granular ball generation method for classification.
\newblock \emph{IEEE Transactions on Emerging Topics in Computational Intelligence}, 8(2): 2022--2036.

\bibitem[{Xu et~al.(2015)Xu, Wang, Tian, Xu, Zhao, Wang, and Hao}]{xu2015short}
Xu, J.~M.; Wang, P.; Tian, G.~H.; Xu, B.; Zhao, J.; Wang, F.~Y.; and Hao, H.~W. 2015.
\newblock Short text clustering via convolutional neural networks.
\newblock In \emph{Proceedings of the Workshop on Vector Space Modeling for Natural Language Processing}, 62--69.

\bibitem[{Yan et~al.(2020)Yan, Fan, Li, Liu, Zhang, Wu, and Lam}]{yan2020unknown}
Yan, G.; Fan, L.; Li, Q.; Liu, H.; Zhang, X.; Wu, X.-M.; and Lam, A.~Y. 2020.
\newblock Unknown intent detection using Gaussian mixture model with an application to zero-shot intent classification.
\newblock In \emph{Proceedings of the Annual Meeting of the Association for Computational Linguistics}, 1050--1060.

\bibitem[{Yang et~al.(2024)Yang, Liu, Xia, Wang, Zhang, Li, and Xu}]{yang20243wc}
Yang, J.; Liu, Z.; Xia, S.; Wang, G.; Zhang, Q.; Li, S.; and Xu, T. 2024.
\newblock 3WC-GBNRS++: A novel three-way classifier with granular-ball neighborhood rough sets based on uncertainty.
\newblock \emph{IEEE Transactions on Fuzzy Systems}, 32(8): 4376--4387.

\bibitem[{Yang et~al.(2022)Yang, Li, Meng, Yang, Liu, and Li}]{yang2022three}
Yang, X.; Li, Y.~J.; Meng, D.; Yang, Y.~X.; Liu, D.; and Li, T.~R. 2022.
\newblock Three-way multi-granularity learning towards open topic classification.
\newblock \emph{Information Sciences}, 585: 41--57.

\bibitem[{Zhan et~al.(2021)Zhan, Liang, Liu, Fan, Wu, and Lam}]{zhan2021out}
Zhan, L.~M.; Liang, H.~W.; Liu, B.; Fan, L.; Wu, X.~M.; and Lam, A.~Y. 2021.
\newblock Out-of-scope intent detection with self-supervision and discriminative training.
\newblock In \emph{Proceedings of the Annual Meeting of the Association for Computational Linguistics and the International Joint Conference on Natural Language Processing}, 3521--3532.

\bibitem[{Zhang, Xu, and Lin(2021)}]{zhang2021deep}
Zhang, H.; Xu, H.; and Lin, T.-E. 2021.
\newblock Deep open intent classification with adaptive decision boundary.
\newblock In \emph{Proceedings of the AAAI Conference on Artificial Intelligence}, 14374--14382.

\bibitem[{Zhang et~al.(2021)Zhang, Xu, Lin, and Lyu}]{zhang2021discovering}
Zhang, H.; Xu, H.; Lin, T.-E.; and Lyu, R. 2021.
\newblock Discovering new intents with deep aligned clustering.
\newblock In \emph{Proceedings of the AAAI Conference on Artificial Intelligence}, 14365--14373.

\bibitem[{Zhang et~al.(2023{\natexlab{a}})Zhang, Xu, Zhao, and Zhou}]{zhang2023learning}
Zhang, H.; Xu, H.; Zhao, S.; and Zhou, Q. 2023{\natexlab{a}}.
\newblock Learning discriminative representations and decision boundaries for open intent detection.
\newblock \emph{IEEE/ACM Transactions on Audio, Speech, and Language Processing}, 31: 1611--1623.

\bibitem[{Zhang et~al.(2023{\natexlab{b}})Zhang, Wu, Xia, Zhao, Gao, Cheng, and Wang}]{zhang2023incremental}
Zhang, Q.; Wu, C.; Xia, S.; Zhao, F.; Gao, M.; Cheng, Y.; and Wang, G. 2023{\natexlab{b}}.
\newblock Incremental learning based on granular ball rough sets for classification in dynamic mixed-type decision system.
\newblock \emph{IEEE Transactions on Knowledge and Data Engineering}, 35(9): 9319--9332.

\bibitem[{Zheng, Chen, and Huang(2020)}]{zheng2020out}
Zheng, Y.~H.; Chen, G.~Y.; and Huang, M.~L. 2020.
\newblock Out-of-domain detection for natural language understanding in dialog systems.
\newblock \emph{IEEE/ACM Transactions on Audio, Speech, and Language Processing}, 28: 1198--1209.

\bibitem[{Zhou, Liu, and Chen(2021)}]{zhou2021contrastive}
Zhou, W.~X.; Liu, F.~Y.; and Chen, M.~H. 2021.
\newblock Contrastive out-of-distribution detection for pretrained transformers.
\newblock In \emph{Proceedings of the Conference on Empirical Methods in Natural Language Processing}, 1100--1111.

\bibitem[{Zhou, Liu, and Qiu(2022)}]{zhou2022knn}
Zhou, Y.; Liu, P.; and Qiu, X. 2022.
\newblock {KNN}-contrastive learning for out-of-domain intent classification.
\newblock In \emph{Proceedings of the Annual Meeting of the Association for Computational Linguistics}, 5129--5141.

\end{thebibliography}

\end{document}